\documentclass[journal]{IEEEtran}
\usepackage{amssymb,amsmath,array,amsfonts, multirow, graphicx,color,url}
\usepackage[latin1]{inputenc}
\usepackage{color}
\usepackage{algorithmic}
\usepackage[ruled,vlined]{algorithm2e} 
\usepackage{bm}
\usepackage{textcomp}

\newcommand{\bA}{\mathbf{A}}
\newcommand{\bB}{\mathbf{B}}
\newcommand{\bO}{\mathbf{0}}

\newcommand{\be}{\mathbf{e}}

\newcommand{\bt}{\mathbf{t}}

\newcommand{\bW}{\mathbf{W}}

\newcommand{\bX}{\mathbf{X}}

\newcommand{\bY}{\mathbf{Y}}
\newcommand{\bz}{\mathbf{z}}

\newcommand{\bst}{\boldsymbol{t}}

\newcommand{\bsy}{\boldsymbol{y}}

\newcommand{\cL}{\mathcal{L}}

\newcommand{\bspi}{\boldsymbol{\pi}}

\newcommand{\bsbeta}{\boldsymbol{\beta}}

\newcommand{\bsSigma}{\boldsymbol{\Sigma}}

\newcommand{\bstheta}{\boldsymbol{\theta}}

\newcommand{\E}{\mathbb{E}}

\newcommand{\R}{\mathbb{R}}
	
\newcommand{\N}{\mathcal{N}}

\newcommand{\bbeta}{\boldsymbol{\beta}}

\begin{document}

\title{An Unsupervised Approach for Automatic Activity Recognition based on Hidden Markov Model Regression}
\author{D.~Trabelsi,
        S.~Mohammed,
        F.~Chamroukhi,
        L.~Oukhellou,
        and~Y.~Amirat 
\thanks{D. Trabelsi, S. Mohammed and Y. Amirat are with University Paris-Est Cr\'eteil (UPEC), LISSI, 122 rue Paul Armangot, 94400, Vitry-Sur-Seine, France e-mail: dorra.trabelsi@etu.u-pec.fr, samer.mohammed@u-pec.fr, amirat@u-pec.fr. 
F. Chamroukhi is with University Sud Toulon-Var, LSIS Lab UMR CNRS 7296, email: Faicel.Chamroukhi@univ-tln.fr. 
L. Oukhellou is with University Paris-Est, IFSTTAR, GRETTIA, F-93166 Noisy-le-Grand, France, e-mail: latifa.oukhellou@ifsttar.fr}
}
\maketitle
\begin{abstract}
Using supervised machine learning approaches to recognize human activities from on-body wearable accelerometers generally requires a large amount of labelled data. When ground truth information is not available, too expensive, time consuming or difficult to collect, one has to rely on unsupervised approaches. This paper presents a new unsupervised approach for human activity recognition from raw acceleration data measured using inertial wearable sensors. The proposed method is based upon joint segmentation of multidimensional time series using a Hidden Markov Model (HMM) in a multiple regression context. The model is learned in an unsupervised framework using the Expectation-Maximization (EM) algorithm where no activity labels are needed. The proposed method takes into account the sequential appearance of the data. It is therefore adapted for the temporal acceleration data to accurately detect the activities. It allows both segmentation and classification of the human activities. Experimental results 
are provided to demonstrate the efficiency of the proposed approach  with respect to standard supervised and unsupervised classification approaches.\\

%

\end{abstract}


\section{Introduction}
\label{sec:Intro}
\IEEEPARstart{T}{he} aging population has recently gained an increasing attention due to its socio-economic impact. By $2050$, the number of people in the European Union aged $65$ and above is expected to grow by $70\%$ and the number of people aged over $80$ by $170\%$\footnote{http://ec.europa.eu/health-eu/my$\_$health/elderly/}. This demographic change poses increasing challenges for healthcare services and their adaptation to the needs of this aging population. Facing this problem or reducing its effect would have a great societal impact by improving the quality of life and regaining people independence to make them active in society. The aim is therefore to facilitate the daily activity lives of elderly or dependent people at home, to increase their autonomy and to improve their safety. In fact, most elderly prefer to stay at home in the so-called ``aging in place" \cite{Kaluza}. The emergence of novel adapted technologies such as wearable and ubiquitous technologies is becoming a privileged solution to provide assistive services to humans, such as health monitoring, well being, security, etc. Among which, activity recognition has a wide range of promising applications in security monitoring as well as human machine interaction \cite{Lu2009}. A large amount of work has been done in this active topic over the past  decades; nevertheless it is still an open and challenging problem \cite{Jinhui}.\\
Several techniques have been used to quantify these activities such as video-based sensors \cite{Brdiczka2009}, wearable-based sensors, environmental sensors and object sensors (smart phones, RFID, etc.). Recently, the use of wearable-sensors based systems for activity recognition has gained more attention on a large number of technological fields such as navigation, monitoring and control of aircrafts \cite{MacKenzie,Carminati}, medical application \cite{Jovanov, Wu}, localization and robots \cite{Barshan, Tan}. Among the inertial sensors used for activity recognition, the accelerometers are the most commonly used \cite{Jiayang}. They have shown satisfactory results to measure the human activities in both laboratory/clinical and free-living environment settings \cite{Mathie}. In addition, the latest advances in Micro-ElectroMechanical Systems (MEMS) technology have greatly promoted the use of accelerometers thanks to the considerable reduction in size, cost and energy consumption. Early studies in activity recognition used uniaxial accelerometers, while recent studies use mainly tri-axial accelerometers \cite{Noury,Yang}.\\ 

\section{Related work on human activity recognition}

Regarding the human activity classification, one can make the distinction between supervised and unsupervised classification approaches. Supervised classification techniques consist in inferring a decision rule from labelled training data. The use of the supervised activity classification approaches has shown promising results \cite{Altun}. Some supervised approaches have enhanced the activity recognition process performances by using spatio-temporal information \cite{Chen}. Regarding the algorithms used in the supervised context, one can cite $k$-Nearest Neighbor ($k$-NN) algorithm \cite{Liu}, multi-class Support Vector Machines (SVM) \cite{Qian} and Artificial Neural Networks (ANN) including both MultiLayer Perceptron (MLP) \cite{Dong,Yang08} and Radial Basis Function (RBF) networks \cite{Faber}. 

Nevertheless, the collection of sufficient amounts of labelled data for a various and rich set of free-living activities may be sometimes difficult to achieve and computationally expensive \cite{Cvetkovic}. 
On the other hand, unsupervised classification techniques try to directly construct models from unlabelled data either by estimating the properties of their underlying probability density (called density estimation) or by discovering groups of similar examples (called clustering).  The unsupervised learning techniques are of particular interest for an exploratory analysis of large amounts of unlabelled data. They can also consist in a preliminary task to further run a supervised classifier based on the obtained partition of the data. The use of an unsupervised approach may be needed in such a context of activity recognition when it is difficult to have labels for the data. 

Regarding the approaches used in the unsupervised context, one can cite the well-known $k$-Means algorithm \cite{Duda}, the Gaussian Mixture Models (GMM) approach \cite{Allen} and the one based on Hidden Markov Model (HMM) \cite{Lin, rabiner} or HMM with GMM emission probabilities \cite{Mannini}. Both the GMM and the HMM approaches use the EM algorithm \cite{dlr}. 

The HMM has shown good results in earlier exploratory studies thanks to their main advantage of suitability to model sequential data which is the case of monitoring human activities. Indeed, the acceleration data are measured over time during physical human activities of a person and are therefore sequential over time. The EM algorithm \cite{dlr} (also called Baum-Welch \cite{BaumWelch}) in the context of HMM is particularity adapted for unsupervised learning.\\

In this study, an unsupervised approach for human activity recognition is proposed. It combines  an HMM-based model with the use of acceleration data acquired during sequences of different human activities. More specifically, the proposed approach is based on a Hidden Markov Model in a multiple regression context and will be denoted by MHMMR.
 
As the sequences of acceleration data consist in multidimensional time series where each dimension is an acceleration, the activity recognition problem is therefore formulated through the proposed MHMMR model as the one of joint segmentation of multidimensional time series, each segment is associated with an activity. 
In the proposed model, each activity is represented by a regression model and the switching from one activity to another is governed by a hidden Markov chain. The MHMMR parameters are learned in an unsupervised way from unlabelled  raw acceleration data acquired during human activities. 

The most likely sequence of activities is then estimated using the Viterbi algorithm \cite{Viterbi}. The proposed technique is then evaluated on real-world acceleration data collected from three sensors placed at the chest, the right thigh and the left ankle of the subject.\\ 
This study is an extension of the paper \cite{Trabelsi2012} where additional technical implementations are shown: Twelve activities and transitions are studied and performances of the proposed approach are evaluated and compared to those of some well known unsupervised and supervised techniques for activity recognition.\\
This paper is organized as follows; section \ref{sec: experimental setup} presents the experimental protocol and the data acquisition platform. Section \ref{sec: Seg with hidden Markov model regression} presents the proposed model and its unsupervised  parameter estimation technique from unlabelled acceleration data. In section \ref{sec: experimental study}, the performances of the proposed approach are evaluated and compared to those of some well known unsupervised and supervised techniques for activity recognition.

\section{Data Collection}
\label{sec: experimental setup}

\begin{figure}[!h]
\centering
\includegraphics[width=6.5cm, height=3.5cm]{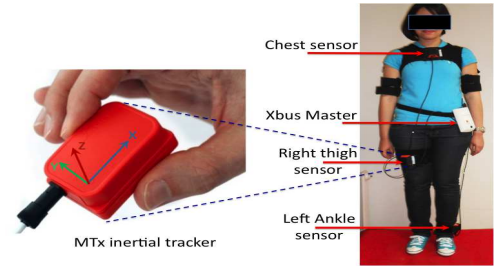}
\caption{MTx-Xbus inertial tracker and sensors placement}
\label{fig_MTX}
\end{figure}

\begin{figure}[!h]
\centering
\includegraphics[width=6.5cm, height=3cm]{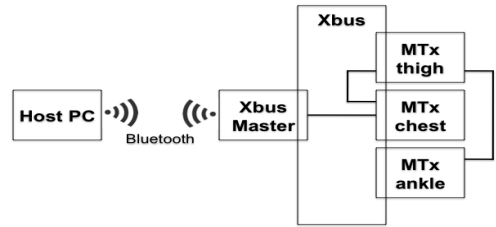}
\caption{Data ghathering from the MTx-Xbus acquisition system}
\label{fig_data_gather}
\end{figure}

In this study, human activities are classified using three sensors placed at the chest, the right thigh and the left ankle respectively as shown in Figure \ref{fig_MTX}. Sensors placement is chosen to represent predominantly upper-body activities such as standing up, sitting down, etc. and predominantly lower body activities such as walking, stair ascent, stair descent, etc. The sensor's placement guarantees at the same time less constraint and better comfort for the wearer. The attachment of the sensors to the human body should be well fitted and secured (Fig. \ref{fig_MTX}). These sensors consist of three MTx 3-DOF inertial trackers developed by Xsens Technologies \cite{Xsens}. Each MTx unit consists of a tri-axial accelerometer measuring the acceleration in the 3-D space (with a dynamic range of $\pm$5g where g represents the gravitational constant). Our experiences show also that the measured ankle-sensor accelerations during the different activities do not  exceed the limit of $\pm$5g. The sampling frequency is set to 25 Hz, which is sufficient and larger than 20 Hz the required frequency to assess daily physical activity \cite{Bouten}. The sensors were fixed on the subject with the help of an assistant before the beginning of the measurement operation. Raw acceleration data are therefore collected over time when performing the activities. The MTx units are connected to a central unit called Xbus Master that is attached to the subject's belt.  Figure \ref{fig_data_gather} shows the data gathering process from the Xbus-MTx acquisition system to the host pc. The Xbus Master is directly connected to the chest MTx unit while the remainig MTx units (thigh and ankle) are connected in series. Data transmission between the Xbus Master and the pc is carried out through a Bluetooth wireless link. 

\begin{figure}[!h]
\centering
\includegraphics[width=8cm, height=3.5cm]{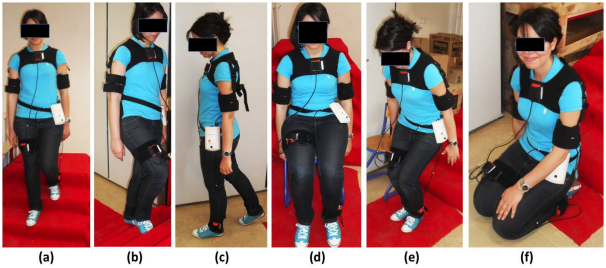}
\caption{Examples of some considered activities: a) Climbing Stairs Down, b) Climbing Stairs Up, c) Walking, d) Sitting, e) Standing Up, f) Sitting on the ground.}
\label{fig:Activities}
\end{figure}

The experiments were performed at the LISSI Lab/University of Paris-Est Cr\'eteil (UPEC) by six different healthy subjects of different ages (who are not the researchers) in the office environment. In order to gather  various and rich dataset, the recruited volunteer subjects have been chosen in a given margin of age (25-30) and weight (55-70) kg. Twelve activities and transitions were studied and are listed as follows: Stairs down (A$1$) - Standing (A$2$) - Sitting down (A$3$) - Sitting (A$4$) - From sitting to sitting on the ground (A$5$) - Sitting on the ground (A$6$) - Lying down (A$7$) - Lying (A$8$) - From lying to sitting on the ground (A$9$) - Standing up (A${10}$) - Walking (A${11}$) - Stairs up (A${12}$). The activities were chosen to have an appropriate representation of everyday activities involving different parts of the body (fig. \ref{fig:Activities}). The recognized activities and transition differ in duration and intensity level. Note that the activities $A_3$, $A_5$, $A_7$, $A_9$ and $A_{10}$ represent dynamic transitions between static activities. Each subject was asked to perform the twelve activities in his own style and was not restricted on how the activities should be performed but only with the sequential activities order. In addition, the duration of each activity is not restricted to be the same as it may vary from one subject to another. 


With three MTx sensor units, each one with a tri-axial accelerometer, a total of nine accelerations are therefore measured and recorded overtime for each activity. 
Since the goal is to recognize human activities from only the raw acceleration data, the acquired acceleration signals can be seen as multidimensional time series (of dimension 9) with regime changes due to the changes of activities over time. The activity recognition problem can therefore be formulated as a problem of multidimensional time series segmentation. Indeed, segmenting the time series according to different unknown regimes over time is equivalent to classifying the acceleration data into one set of activities; each activity being associated with a regime. This will be detailed in the next section that is dedicated to the proposed  Hidden Markov Model Regression (HMMR) approach. 

\section{Segmentation with Multiple Hidden Markov Model Regression - MHMMR } 
\label{sec: Seg with hidden Markov model regression}
In this section, the problem of activity recognition (classification) is formulated as the one of joint segmentation of multidimensional time series. 
Indeed, the acceleration data are presented as multidimensional time series presenting various regime changes. In such context, the goal is to provide an automatic partition of the data into different segments (regimes), each segment being considered afterwards as an activity. 

Various modelling approaches of time series presenting regime changes have been proposed in literature. One can cite in particular the piecewise regression as one of the most adapted modelling approaches \cite{McGee,Brailovsky}. The piecewise model has been applied in many domains including finance, engineering, economics, and bioinformatics \cite{Picard}. In the piecewise regression model \cite{Brailovsky}, data are  partitioned into several segments, each segment being characterized by its mean polynomial curve and its variance. However, the parameter estimation in such method requires the use of dynamic programming algorithm \cite{Bellman,Stone} which may be computationally expensive especially for time series with large number of observations. Moreover, the standard piecewise regression model usually assumes that noise variance is uniform for all the segments (homoskedastic model). An alternative approach extended in this paper is based on Hidden Markov Model Regression  \cite{Fridman}. 
This approach can be seen as an extension of the standard Hidden Markov Model (HMM) \cite{rabiner} to regression analysis. Each regime is described by a regression model rather than a simple constant mean over time, while preserving the Markov process modelling for the sequence of unknown (hidden) activities. Indeed, standard HMM-based approaches use simple Gaussian densities as density of observation. However, in the HMM regression context, each observation is assumed to be a noisy  polynomial function to better model very structured data as the acceleration data. The approach we propose further extends the HMM model to a multiple regression setting. This is due to the fact the observed acceleration data is multidimensional. 
In the following, the Hidden Markov Regression Model for time series modelling is used by formulating its basic and multiple regression setting. In this framework, each observation, denoted by $\bsy_i$, represents the $i$th acceleration measurement while the associated state (class), denoted by $z_i$, represents its corresponding activity.

\subsection{General description of the Multiple Hidden Markov Model  Regression}
\label{ssec: MHMMR}

In Hidden Markov Model Regression (HMMR), each time series is represented as a sequence of observed univariate variables $(y_1, y_2, \hdots, y_n)$, where the observation $y_i$ at time $t_i$ is assumed to be generated by the following regression model \cite{Fridman}:
\begin{equation}
y_i = \bsbeta^T_{z_i}\bt_i + \sigma_{z_i}\epsilon_i \quad ; \quad\epsilon_i \sim \N(0,1) , \quad (i=1,\ldots,n)
\label{eq: HMM regression model}
\end{equation}
where $z_i \in \{1,\ldots,K\}$ is a hidden discrete-valued variable. In this application case,  $z_i$ represents the hidden class label (activity) of each acceleration data point and $K$ corresponds to the number of considered activities. The variable $z_i$ controls the switching from one polynomial regression model associated to one activity, to another of $K$ models at time $t_i$. The vector $\bsbeta_{z_i}=(\beta_{z_i0},\ldots,\beta_{z_i p})^T$ is the one of regression coefficients of the $p$-order polynomial regression model $z_i$ and $\sigma_{z_i}$ is its corresponding standard deviation, $\bt_i=(1, t_i,t_i^{2} \ldots, t_i^{p})^T$ is the $p+1$ dimensional covariate vector at time $t_i$ and the $\epsilon_i$'s are standard Gaussian variables representing an additive noise. The HMMR assumes that the hidden sequence $\bz=(z_1,\ldots,z_n)$ is a homogeneous Markov chain of first order parameterized by the initial state distribution $\pi$ and the transition matrix $\bA$. It
can be shown that, conditionally on  a regression model $k$ ($z_i=k$), $y_i$ has a Gaussian distribution with mean $\bbeta_k^T\bst_i$ and variance $\sigma_k^2$. Regarding the multiple regression case, the model can be formulated as follows:
\begin{eqnarray}
y^{(1)}_i &=& \bsbeta^{(1)T}_{z_i}\bt_i + \sigma^{(1)}_{z_i}\epsilon_i \nonumber \\
y^{(2)}_i &=& \bsbeta^{(2)T}_{z_i}\bt_i + \sigma^{(2)}_{z_i}\epsilon_i \nonumber \\
\vdots & & \vdots \nonumber \\
y^{(d)}_i &=& \bsbeta^{(d)T}_{z_i}\bt_i + \sigma^{(d)}_{z_i}\epsilon_i 
\label{eq: M-HMM regression model}
\end{eqnarray}
where $d$ represents the dimension of the time series (sequence) and the latent process $\bz$ simultaneously governs all the univariate time series components. 
The model (\ref{eq: M-HMM regression model}) can be rewritten in a matrix form as follows:
 \begin{equation}
\bsy_i =  \bB_{z_i}^T  \bt_i + \be_i \quad ; \quad\be_i \sim \N(\bO,\bsSigma_{z_i}) , \quad (i=1,\ldots,n)
\label{eq: M-HMM regression model matrix form}
\end{equation}
where $\bsy_i=(y^{(1)}_i,\ldots,y^{(d)}_i)^T$ is the $i$th observation of the time series in $\R^d$, $\bB_k = \left[\bsbeta^{(1)}_{k},\ldots, \bsbeta^{(d)}_{k} \right]$ is a $(p+1)\times d$ dimensional matrix of the  multiple regression model parameters associated with the regime (class) $z_i=k$ and $\bsSigma_{z_i}$ 
its corresponding covariance matrix.

The Multiple HMMR model is therefore fully parameterized by the parameter vector $\bstheta=(\bspi,\bA,\bB_1,\ldots,\bB_K,\bsSigma_1,\ldots,\bsSigma_K)$. The next sub-section gives the parameter estimation technique by maximizing the observed data likelihood through the Expectation-Maximization (EM) algorithm.
The parameter vector $\bstheta$ is estimated using the well-known maximum likelihood
method thanks to its very well-known attractive limiting properties of consistency, asymptotic normality and efficiency. Indeed, in our experiments, a considerable number of data points is acquired during time, which makes the sample size suitable to take advantage of the limiting properties of the maximum likelihood estimator. 
The log-likelihood to be maximized in this case is written
as follows: 
{\small{
\begin{eqnarray}
\!\!\!\!\!\!\!\!\cL(\bm{\theta})&\!\!\!\!\!\!\!\!=\!\!\!\!\!\!\!\!&\log p({\bm{y}_1},\ldots,{\bm{y}_n};\bm{\theta}) \nonumber\\
\!\!\!\!\!\!\!\!&\!\!\!\!\!\!=\!\!\!& \!\!\log\sum_{\bz} p(z_1;\pi)\prod_{i=2}^{n}p(z_i|z_{i-1};A)\prod_{i=1}^{n}\mathcal{N}({\bm{y}_i};\bB^T_{z_i}\bt_i,{\Sigma_{z_i}})
\label{HMM_regression_logik}
\end{eqnarray}
}
}
Since this log-likelihood cannot be maximized directly, this can be performed using the EM algorithm  \cite{dlr,McLachlan},  that  is known as the Baum-Welch algorithm in the HMM context \cite{BaumWelch,rabiner}. This algorithm alternates between the two following steps:

\paragraph{E-step} This step computes the conditional expectation of the complete-data log-likelihood given the observed data $\bY$, time $\bt$ and a current parameter estimation denoted by $\bstheta^{(q)}$:
\begin{equation}
Q(\bstheta,\bstheta^{(q)})= \E\Big[\log p(\bY,\bz|\bt;\bstheta)|\bY,\bt;\bstheta^{(q)}\Big]\cdot
\label{eq: Q-function for M-HMMR}
\end{equation}
It can be easily shown that this step only requires the calculation of:  
\begin{itemize} \item the posterior probability 
\begin{equation}
\tau^{(q)}_{ik} \!\!= p(z_i=k|\bY,\bt;{\bm{\theta}^{(q)}})
\label{eq:the posterior probability}
\end{equation}
$\forall$ $i=1,\ldots,n$ and $k=1,\ldots,K$ which is the posterior probability that $\bsy_i$ originates from the $k$th polynomial regression model  given the whole  observation sequence and the current parameter estimation $\bstheta^{(q)}$,
\item and the joint posterior probability of the state $k$ at time $i$ and the state $\ell$ at time $i-1$ given the whole observation sequence and the current parameter estimation $\bstheta^{(q)}$, that is 
\begin{equation}
 \xi^{(q)}_{i\ell k} \!\! = \!\!p(z_{i}=k, z_{i-1}=\ell|\bY,\bt;{\bm{\theta}^{(q)}}) 
\label{eq:joint posterior probability}
\end{equation}
$\forall$ $i = 2,\ldots,n$ and $k,\ell=1,\ldots,K$.
\end{itemize}

These posterior probabilities are computed by the forward-backward procedures in the same way as for a standard HMM \cite{rabiner}. More calculation details on this step can be found in \cite{rabiner}. 

\paragraph{M-step} In this step, the value of the parameter  $\bstheta$ is updated by computing the parameter $\bstheta^{(q+1)}$ that maximizes the conditional expectation (\ref{eq: Q-function for M-HMMR}) with respect to $\bstheta$. It can be shown that this maximization leads to the following updating rules. 
The updates of the parameters governing the hidden Markov chain $\bz$ are the ones of a standard HMM and are given by:
\begin{equation}
\pi^{(q+1)}_k = \tau^{(q)}_{1k} 
\label{eq: updating pi for the HMC}
\end{equation}
\begin{equation}
\bA^{(q+1)}_{\ell k} = \frac{\sum_{i=2}^{n}\xi^{(q)}_{ik\ell}}{\sum_{i=2}^{n}\tau^{(q)}_{ik}} 
\label{eq: updating A for the HMC}
\end{equation}
Updating the regression parameter consists in performing $K$ weighted multiple polynomial regressions. The regression parameter matrices updates are given by :

\begin{eqnarray}
{\bB}_k^{(q+1)} &=& \Big[\sum_{i=1}^{n}\tau^{(q)}_{ik} \bt_i\bt_i^T \Big]^{-1} \sum_{i=1}^{n}\tau^{(q)}_{ik} \bt_i \bsy^T_i   \nonumber \\
&=& (\bX^T\bW_k^{(q)}\bX)^{-1}\bX^T\bW_k^{(q)}\bY,
\label{eq: EM estimate of B_k for the MHMMR}
\end{eqnarray}
where $\bW_k^{(q)}$ is a $n \times n$ diagonal matrix of weights whose diagonal elements are the posterior probabilities $(\tau_{1k}^{(q)},\ldots,\tau_{nk}^{(q)})$ and $\bX$ is the $n\times (p+1)$ regression matrix given by $(\bt_1,\ldots,\bt_n)^T$.
The updating rule for the covariance matrices is written as a weighted variant of the estimation of a multivariate Gaussian density with the polynomial mean ${\bB}_k^{T(q+1)}\bt_i$ such as:
{\small{
\begin{eqnarray}
{\bsSigma}_k^{(q+1)}\!\!\!\!\!\!&=&\!\!\!\!\!\!\frac{1}{\sum_{i=1}^{n} \tau^{(q)}_{ik}}\sum_{i=1}^{n} \tau^{(q)}_{ik} (\bsy_i-{\bB}_k^{T(q+1)}\bt_i)^T(\bsy_i-{\bB}_k^{T(q+1)}\bt_i) \nonumber\\
\!\!\!\!\!\!&=&\!\!\!\!\!\! \frac{1}{\sum_{i=1}^{n} \tau^{(q)}_{ik}}(\bY - \bX \bB_k^{(q+1)})^T \bW_k^{(q)} (\bY - \bX \bB_k^{(q+1)})
\label{eq: EM estimate of Sigma_k for the MHMMR}
\end{eqnarray}
}
}

\section{Results and discussions}
\label{sec: experimental study}

This section presents experiments carried out to validate the two main ideas explored  throughout this paper, i.e., the segmentation and the classification of the human activity from raw acceleration data using a MHMMR approach within an unsupervised learning framework\footnote{Note that, in this study, the raw acceleration data are directly used without any feature extraction. Indeed, in many area of application a feature extraction step is needed before running the classifier and may itself lead to an additional computational cost, which can be penalizing in real time applications.}. Series of experiments were conducted to evaluate the performance of the proposed approach and also to perform comparisons with well-known unsupervised and supervised classification approaches.

\subsection{Performance evaluation}
\label{sec: Performance evaluation}
Given a set of 9-dimensional acceleration data from three triaxial accelerometer modules mounted on the chest, right thigh and left ankle, the proposed approach allows both segmentation and classification of the twelve activities. Each obtained segment is indeed considered as an activity, achieving thus a classification task. We chose to take as a ground truth about the class of an activity, labeling obtained thanks to an expert. While the different subjects were performing the sequence of activities, an independent operator was asked to annotate the activities, thus providing a labeling of the dataset\footnote{Note that the labels were not used to train unsupervised models; they were only used afterwards for the evaluation of classification errors.}. The provided partition is indeed matched to the true labels (ground truth) by evaluating all the possible label switchings. The label switching leading to the minimum error rate is selected as the best class prediction. For the supervised classification approaches, data labels were used to both train and test the models. In this case, the performance was estimated through a 10-fold cross-validation procedure.
Regarding the classification problem, confusion matrices between the annotated classes and the estimated classes for all the subjects in the database are computed. The criteria used to evaluate the performance of an approach are the correct classification rate and the prediction accuracy in terms of precision and recall.
%
%
%

In the following, the results of the MHMMR approach obtained on real acceleration data of human activities are first detailed, then they are compared to those of standard unsupervised and supervised classification approaches.

\subsection{Classification performance of the MHMMR}

\begin{figure}[!t]
\centering
\includegraphics[width=8cm, height=6cm]{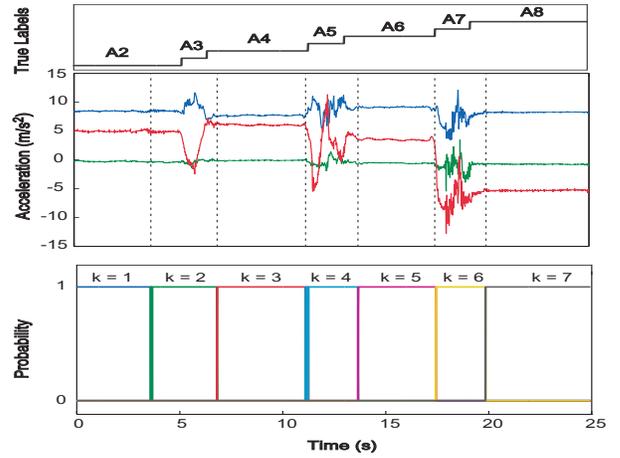}
\caption{MHMRM segmentation for the sequence (Standing A$2$ - Sitting down A$3$ - Sitting A$4$ - From sitting to sitting on the ground A$5$ - Sitting on the ground A$6$ - Lying down A$7$- Lying A$8$)  for the seven classes k=(1,\ldots, 7)}\label{fig:sequence1}
\end{figure}

\begin{figure}[!t]
\centering
\includegraphics[width=8cm, height=6cm]{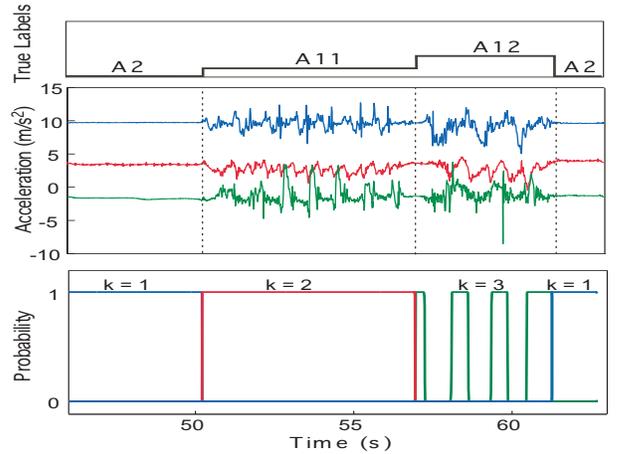}
\caption{MHMMR segmentation for the sequence (Standing A$2$ - Walking A${11}$ - Climbing up stairs A${12}$ - Standing A$2$) for the three classes k= {1,\ldots ,3}}\label{fig:sequence2}
\end{figure}

The following experiments were conducted to qualitatively assess the performances of the proposed approach in terms of automatic segmentation of human activity on the basis of raw acceleration signals. From the sequence of nine observed variables $\bsy_i = (y^{(1)}_i,\dots,y^{(9)}_i)$ at each time step $i$ for $i=1,\ldots,n$ corresponding to the 3-axis accelerations measured by the three sensors, the MHMMR is used to identify the latent sequence $\bz = (z_1, \dots, z_n)$ corresponding to the twelve activities. The number of classes $K$ is fixed to twelve and the order of regression $p$ is fixed empirically to three as it gives the best performance among several values of $p$. Model parameters are estimated from the data using the algorithm detailed in Section \ref{ssec: MHMMR}. Figures \ref{fig:sequence1}  and \ref{fig:sequence2} show the performance of the proposed method to segment the two following sequences: 
\begin{itemize}
\item Sequence 1: Standing - Sitting down - Sitting - From sitting to sitting on the ground - Sitting on the ground - Lying down  - Lying, 
\item Sequence 2: Standing  - Walking - Climbing up stairs  - Standing. 
\end{itemize}
These figures represent the evolution of the acceleration data and the corresponding posterior probabilities for the two different sequences. Note that the posterior probability is the probability that a sample $i$ will be generated by the regression model $k$ given the whole sequence of observations $(\bsy_1,\dots,\bsy_n)$. It can be observed that the obtained sequences are interesting and promising despite some confusion between activities such as (A${11}$, A${12}$).

Table \ref{table:chest-thigh-right} shows that the MHMMR gives $91.4\%$ as a mean correct classification rate averaged over all observations. It highlights the potential benefit of the proposed approach in terms of automatic segmentation and classification of human activity. Both the transitions and the stationary activities are well identified. Exhaustively, table \ref{table:Prec-recal-Act} gives the percentage of precision and recall for each activity. 
Indeed, one can observe that static activities (A${2}$, A${4}$, A${6}$ and A${7}$) are easier to recognize than dynamic activities (A$1$, A${11}$, A${12}$).
\begin{table*}[t!]
\centering
\begin{tabular}{|*{1}{l|} c| c| c| c| c| c| c| c| c| c| c| c|}
   \hline
    Class & A$1$ & A$2$ & A$3$ & A$4$ & A$5$ & A$6$ & A$7$ & A$8$ & A$9$ & A${10}$ & A${11}$ & A${12}$\\
   \hline
  Precision ($\%$)&71.9 &96.4 &78.4 &95.7 &92.3 &98.9 &97.6 &92.5 &82.6 &82.6 & 83.2 &95.6\\
   \hline
    Recall ($\%$) &95 &87.8 &83.4 & 94 &97.3 &94.6 &95.4 &90.9 &98.5 &92.2 & 98.1 &82.3\\
   \hline
\end{tabular}
\caption{Recall versus precision of the MHMMR}
\label{table:Prec-recal-Act}
\end{table*}
In order to focus on the efficiency ratio of the three sensors used for activity recognition, the MHMMR algorithm has been evaluated using data from only two sensors. The classification results, given in Table \ref{table:chest-thigh-right}, show as expected, that the percentage of correctly classified instances decreases with the number of data sources. The worst result is obtained when the sensor placed at the thigh is not taken into account.
\begin{table}[!t]
\centering
\begin{tabular}{|l| c| }
   \hline
   Sensors &  Percentage of correctly classified instances\\
   \hline
   \small {Chest, thigh, ankle} &   $91.4\% \pm 1.65$ \\   
   \hline
   \small {Chest, ankle} &  $83.9\% \pm 1.98$  \\
   \hline
   \small {Chest, thigh} &  $86.2\% \pm 2.03$   \\
   \hline
   \small {Thigh, ankle} & $84\% \pm 2.21$    \\
   \hline
   \end{tabular}
\caption{Effects of reducing the number of sensors when using MHMMR}
\label{table:chest-thigh-right}
\end{table}
\subsection{Comparison with unsupervised and supervised classification approaches}
Correct classification rates and the standard deviations obtained with standard unsupervised and supervised classification approaches as well as the MHMMR approach are given in Table \ref{table:class-prec-recall1}  and Table \ref{table:class-prec-recall2}.
\begin{table}[!h]
\begin{tabular}{|*{1}{l|} c|c| c|}
   \hline
    &   Correct Classification ($\%$)& Precision ($\%$) & Recall ($\%$) \\
   \hline
   $k$-Means  & $60.2 \pm 2.48$ & $60.4$ & $59.8$ \\
   GMM & $72.3 \pm 2.05$ & $71.8$ & $73.5$ \\
   HMM & $84.1 \pm 1.84$ & $83.8$ &$84$  \\
   MHMMR & $91.4 \pm 1.65$ & $89$ &$95.6$  \\
   \hline
\end{tabular}
\caption{Comparison of the performance in terms of Correct Classification, Recall and Precision of the four unsupervised classifiers}
\label{table:class-prec-recall1} 
\end{table}
Compared to standard unsupervised classifiers, the proposed MHMMR outperforms them since it provides a classification rate of 91.4 $\%$ while only 60 $\%$, $72\%$ and $84\%$ of instances are well classified with respectively the $k$-Means, the GMM and the standard HMM approaches. Notice that, the GMM and K-means approaches are not well suitable for this kind of longitudinal data.
\begin{table}[!h]
\centering
\begin{tabular}{|*{1}{l|} c|c| c|}
   \hline
    &   Correct Classification & Precision & Recall \\
    &   ($\%$) & ($\%$) & ($\%$) \\
   \hline
   Naive Bayes  & $80.6 \pm 0.91$ & $80.9$ & $80.6$ \\
   MLP & $83.1 \pm 0.45$ & $82.8$ & $83.2$ \\
   SVM & $88.1 \pm 1.32$ & $87.6$ & $88.3$ \\
  $k$-NN  & $95.8 \pm 0.32$ & $95.9$ & $95.9$ \\
   Random Forest & $93.5 \pm 0.78$ & $93.5$ & $93.5$ \\ 
  \hline
\end{tabular}
\caption{Comparison of the performance in terms of Correct Classification, Recall and Precision of the five supervised classifiers}
\label{table:class-prec-recall2} 
\end{table}
In Table \ref{table:class-prec-recall2}, it can be observed that the $k$-NN ($k=1$) gives the highest classification rates with 95.8$\%$, followed by the Random Forest with 93.5$\%$. Then, the SVM gives 88.1$\%$ and the MLP gives 83.1$\%$. However, the Naive Bayes gives the lowest classification rate with 80.6$\%$. Table \ref{table:class-prec-recall2} shows also that the $k$-NN ($k=1$) has the best classification algorithm in terms of prediction accuracy since it achieves 95.9$\%$ of precision and recall.
Compared to standard supervised classification techniques (using class labels), these results are very encouraging since the proposed approach performs in an unsupervised way. The main errors are due to the confusions located in transition segments.  This is due to the fact that the transitions lengths are much shorter than the activities ones. Since the confusion matrix was computed using real labels supplied by a human expert, the obtained labels may not correspond perfectly to the expert labels, particularly, during transitions. Indeed, it is difficult to have the ground truth of the limit between an activity and a transition. Furthermore, the  aforecited supervised classification approaches  require a labelled collection of data to be trained. Besides, they do not explicit the temporal dependence in their model formulation as they assume an independent hypothesis for the data; the data are treated as several realizations in the multidimensional space ($\R^d$) without considering possible dependencies between the activities. Moreover, it can be noticed that assigning a new sample to a class using the $k$-NN approach requires the computation of as many distances as  there are examples in the dataset, which may lead to a significant computation time. Using the proposed approach, classification needs the computation of the posterior  probabilities, as many as there are activities. On the other hand, comparison with the unsupervised classification approaches ($k$-Means and the GMM) and the standard HMM shows that the proposed method gives relatively a high rate and better performances.

\section{Conclusion and future works}
In this paper, we presented a statistical approach based on hidden Markov models in a regression context for the joint segmentation of multivariate time series of human activities. 
It is based upon the use of raw accelerometer data acquired from body mounted inertial sensors in a  health-monitoring context. The main advantage of the proposed approach comes from the fact that the statistical model explains the regime changes over time through the hidden Markov chain, each regime being interpreted as an activity (a class). Furthermore, learning with this statistical model is performed in an unsupervised way using unlabelled examples only; parameter estimates are computed by maximizing a likelihood criterion, using a dedicated EM algorithm. Considering human activity recognition within an unsupervised learning framework can be particularly interesting within an exploratory data-mining context in order to automatically cluster a large amount of unlabelled acceleration data into different groups of activity. The comparison with well-known supervised classification approaches shows that the proposed method is competitive even when performed in an unsupervised way. This work can be extended in several directions, namely integrating the model into a Bayesian context to better control the model complexity via choosing suitable prior distributions on the models parameters. Then, and perhaps more interestingly, another step to explore is to built a fully non Bayesian non-parametric model which will be useful for any kind of complex activities and in which the number of activities will not have to be fixed.  In terms of application, 
 a promising perspective in a rehabilitation context would be to use the proposed approach for recognizing in an unsupervised framework the undesirable compensatory physical behaviours observed with stroke and injury patients.

\end{document}